\documentclass[runningheads]{llncs}
\usepackage{graphicx}
\usepackage{amsmath,amssymb} 
\usepackage{color}
\usepackage{bm}
\usepackage{cite}
\usepackage[width=122mm,left=12mm,paperwidth=146mm,height=193mm,top=12mm,paperheight=217mm]{geometry}
\usepackage{ulem}
\usepackage{color}

\newcommand{\figref}[1]{Fig.~\ref{#1}}
\newcommand{\tabref}[1]{Table~\ref{#1}}

\newif\iffigure
\figuretrue

\newif\iftable
\tablefalse

\begin{document}
\pagestyle{headings}
\mainmatter
\def\ECCV18SubNumber{12}  

\title{Understanding Fake Faces}


\authorrunning{R. Natsume et al.}
\author{Ryota Natsume$^{1*}$ \and Kazuki Inoue$^{1*}$ \and Yoshihiro Fukuhara$^{1}$ \and Shintaro Yamamoto$^{1}$ \and Shigeo Morishima$^{1}$ \and Hirokatsu Kataoka$^{2}$}
\institute{Waseda University \and National Institute of Advanced Industrial Science and Technology (AIST) \\ ~ \\
\email{nano.poteto@toki.waseda.jp},~ \email{sogew3@ruri.waseda.jp},~ \email{f\_yoshi@ruri.waseda.jp} \\ \email{s.yamamoto@fuji.waseda.jp},~ \email{shigeo@waseda.jp},~ \email{hirokatsu.kataoka@aist.go.jp}}

\maketitle
\begin{abstract}
Face recognition research is one of the most active topics in computer vision (CV), and deep neural networks (DNN) are now filling the gap between human-level and computer-driven performance levels in face verification algorithms. However, although the performance gap appears to be narrowing in terms of accuracy-based expectations, a curious question has arisen; specifically, \textit{Face understanding of AI is really close to that of human?} In the present study, in an effort to confirm the brain-driven concept, we conduct image-based detection, classification, and generation using an in-house created fake face database. This database has two configurations: (i) false positive face detections produced using both the Viola Jones (VJ) method and convolutional neural networks (CNN), and (ii) simulacra that have fundamental characteristics that resemble faces but are completely artificial. The results show a level of suggestive knowledge that indicates the continuing existence of a gap between the capabilities of recent vision-based face recognition algorithms and human-level performance. On a positive note, however, we have obtained knowledge that will advance the progress of face-understanding models.

\end{abstract}

\footnotetext{∗indicates equal contribution}
\section{Introduction}
In the field of computer vision (CV), research on human faces, which includes face detection~\cite{viola01,bai18}, three-dimensional (3D) face reconstruction from images~\cite{blanz99,tran18}, and face recognition~\cite{ahonen06,zhao18} is one of the most active topics. Assisted by the rise of deep neural networks (DNN), vision-based approaches have improved to the point where face verification with DeepFace~\cite{taigman14} and face recognition with FaceNet~\cite{schroff15} now exceed human performance levels.

However, a curious question has arisen; specifically; \textit{``Does artificial intelligence (AI) recognize faces the same way humans do?"} For example, vision-based approaches still have some mistaken case that humans don't have.  (see \figref{fig:concept})

Herein, we consider recent vision-based approaches to human-like face understanding in terms of the two following aspects:
\begin{enumerate}
    \item False-positive face analysis (\figref{fig:faces} (b)(c)): 
    False-positive human face detections are far more likely with an AI face detector than during human observation, but observations in feature space seem to be similar. Hence, face false-positive detections by representative face detection algorithms can help us gain a better grasp of AI face understanding.
    \item Simulacra/pareidolia face analysis (\figref{fig:faces} (d)):
    Simulacra~\cite{baudrillards93} and pareidolia~\cite{liu14} are psychological phenomena that allow humans to recognize particular objects (such as an arrangement of three points resembling two eyes and a mouth) as faces. In other words, simulacra/pareidolia face detections are false positives triggered by human psychological peculiarities.
\end{enumerate}
The analysis of false positives (\figref{fig:faces} (b) and (c)) and simulacra faces (\figref{fig:faces} (d)) may help us form a perspective concerning human-like face recognition. We define the above-mentioned two aspects as \textit{fake faces}.

In this paper, we confirm human-like face recognition parameters by analyzing fake faces. To carry out our experiments, we collected a fake face database that contains (i) false-positive faces extracted via the Viola Jones method~(VJ)~\cite{viola01} and convolutional neural networks~(CNN)~\cite{liu16}, and (ii) simulacra/pareidolia faces. Since we believe that image classification and generation are required to understand fake faces, we will attempt to implement face classification with CNN and conduct fake face generation with generative adversarial networks (GAN)~\cite{goodfellow14} using our fake face database.

In face classification, we begin by training CNN to classify fake faces in order to verify the accuracy of real faces; whereas in fake face generation, we train GAN with the fake face database and determine whether the generated images will be recognized as faces by human observers. The results show a level of suggestive knowledge that indicates the continuing existence of a gap between the capabilities of recent vision-based face recognition algorithms and human-level performance. On a positive note, we have obtained knowledge that will advance the progress of current face understanding models.

\iffigure
\begin{figure}[t]
  \begin{center}
    \includegraphics[clip,scale=0.32]{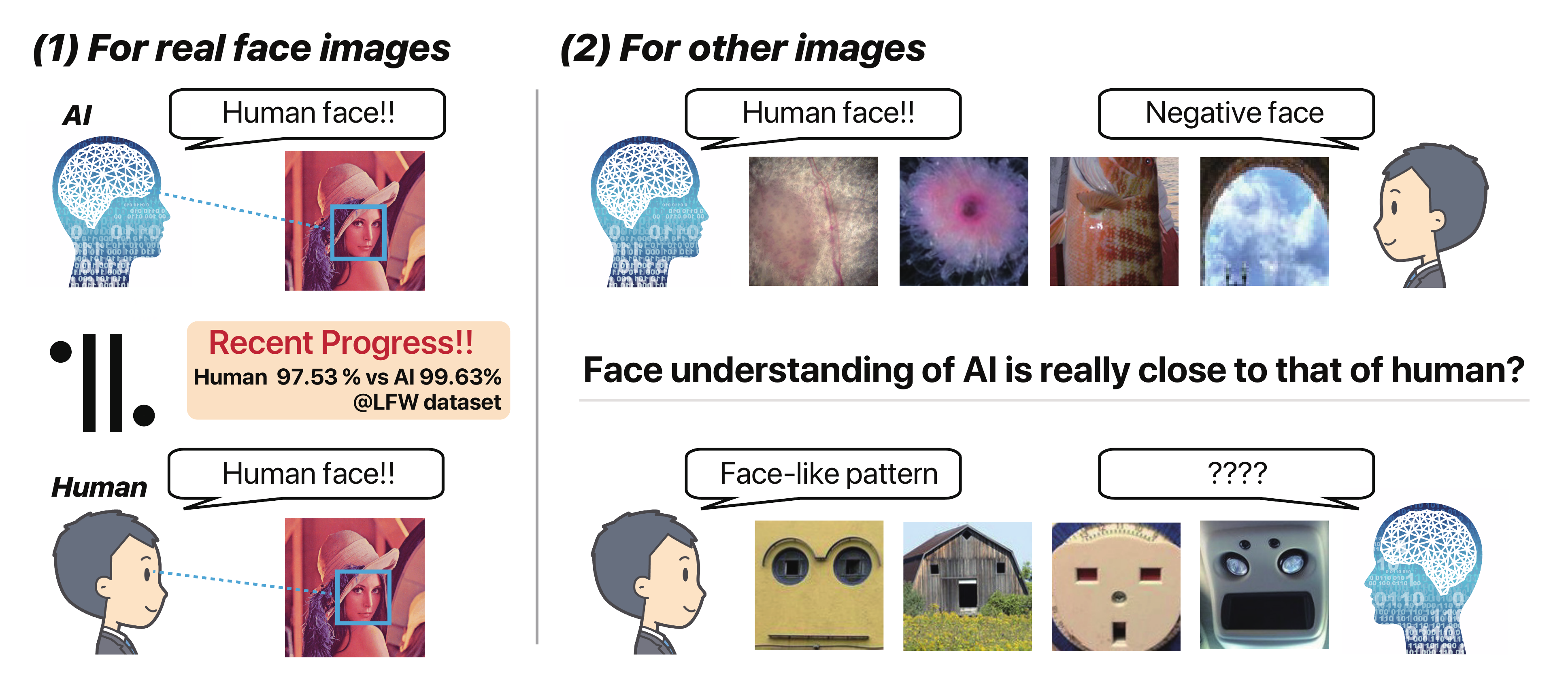}
    \vspace{-10.0pt}\caption{The recent progress (left) and our curious scenario(right) in face understanding: (left) For real face images, we have achieved significant progress with AI-based systems such as DeepFace~\cite{taigman14} and FaceNet~\cite{schroff15}. This is especially notable in FaceNet, which outperformed human-level accuracy by 99.63 to 97.53 on the (top-right) Labeled Faces in the Wild (LFW) dataset. For other images, we found that AI-based systems misinterpret some objects as faces even though humans (bottom-right) can correctly identify negative faces. Humans also correctly recognize face-like objects that are described as simulacra faces. In this paper, we verify the curious scenario  with CV tools such as CNN image classification and GAN image generation.} 
\label{fig:concept}
\end{center}
\end{figure}
\fi

The main contributions of this work include:

\textbf{\underline{Conceptual contribution}:} We confirmed the answer to the question, \textit{``Is AI face understanding actually close to human-level performance?"}, by analyzing performance levels with fake faces. The results of our experiments show that CNN-based approaches have limitations when recognizing human-like faces, and it is thought that a new perspective of joint-understanding with real and fake faces will help facilitate more human-like face understanding.

\textbf{\underline{Database contribution}:} We also present a novel database, referred to as the fake face database, which contains false positives produced by face detector and simulacra/pareidolia faces. The use of this database helped us to confirm the gap between human- and computer-based face understanding.

\iffigure
\begin{figure}[t]
  \begin{center}
    \includegraphics[clip,scale=0.40]{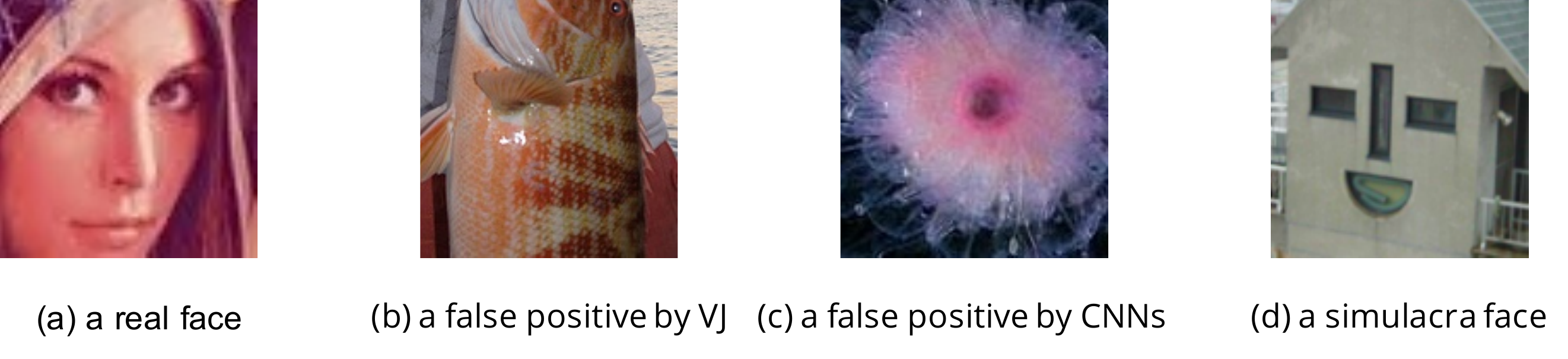}
    \vspace{-15.0pt}\caption{(a) is a real face, (b) and (c) are false positives identified by a hand-crafted face detector (VJ) \cite{viola01} and a CNN face detector \cite{liu16}, respectively; whereas (d) consists of simulacra faces, which are false positives produced by human psychological face recognition characteristics. 
    In this work, we refer to (b), (c) and (d) as fake faces, and examine the relationship between fake faces and real faces (a). \vspace{-13.0pt} \label{fig:faces}} 
  \end{center}
\end{figure}
\fi

\section{Fake faces}
\vspace{-3.0pt}
\subsection{VJ/CNN fake faces}
Face detection research has a long history \cite{zafeiriou15}. Its fundamental approaches are based on shallow learning involving tools such as support vector machines (SVMs) with handcrafted features \cite{viola01}. In recent years, object detection has improved along with the recent progress of DNN algorithms \cite{liu16}. However, neither of these methods have yet achieved $100\% $ accuracy, which means they sometimes detect non face objects, i.e. false positives. As shown in \figref{fig:faces}, false positive faces can be totally unlike real faces. \figref{fig:faces} (b) and (c) show false positives produced by a handcrafted features-based method (VJ) \cite{viola01} and a DNN-based method (CNN face detector), \cite{liu16} respectively. Input images are grayscale in the former and RGB in the latter. Our analysis of these false positive images is expected to help us to understand how AI recognizes faces.

In fact, each of the false positive types have different characteristics because VJ detects faces based on handcrafted features, whereas CNN face detectors detect faces based on millions of learning features. Therefore, in this work, false positives detected by handcrafted features and those detected by deep learning are handled differently. In the next section, we will analyze these two false positive types in an effort to fill the gap between human and AI characteristics. Hereafter, we will refer to VJ and CNN false positives as VJ fake faces and CNN fake faces, respectively.

\subsection{Simulacra Fake faces}

In this paper, simulacra \cite{baudrillards93} refer to false positive recognition triggered by the psychological phenomena of human brains that cause us to perceive three points (arranged appropriately) as a face, as shown in \figref{fig:faces} (d). One proposed theory posits that human brain face recognition is guided primarily by the identification of two eyes and a mouth. Therefore, in spite of possessing wholly unrealistic texture and shapes, a group of three points located on reversed triangle vertices can trigger human false positive face detections.

Pareidolia \cite{liu14} is another of the psychological phenomena that causes humans to recognize wall stains or tree bark patterns as faces. Pareidolia faces also have textures that are associated with two eyes and a mouth, much the same as simulacra faces. In this paper, we refer to both simulacra and pareidolia faces as simulacra fake faces. In \figref{fig:faces}, it can be seen that, in terms of human perception, simulacra fake faces are more identifiable as faces than VJ/CNN fake faces.

However, previous studies~\cite{takahashi15, abaci15} have shown that since handcrafted feature-based face detectors cannot recognize simulacra faces, additional algorithms would be necessary if it were desirable to detect simulacra faces in addition to real faces. Since this study suggests that simulacra faces are not sufficiently similar to human faces to be detected by AI algorithms, we will analyze simulacra fake faces in order to reveal how much vision-based face understanding resembles human perception.

\section{Approaches to Verify computer face understanding }

Next, we analyze VJ, CNN, and simulacra fake faces to verify whether AI algorithms have human-like face understanding. To accomplish this, we conducted three experiments: face detection of simulacra faces, face classification trained with fake faces, and fake-face generation. In simulacra fake-face detection, we examined simulacra false positives to gain an understanding of the gap between humans and AI in face recognition characteristics.
In face classification with fake faces, we determine how similar AI false positives are to real faces. In addition, we confirm whether an AI can learn real face characteristics from simulacra fake faces. In fake face generation, we compare images generated by GAN trained with each fake face type in order gain an understanding of fake face characteristics. The overall goal of these experiments is to clarify the gap between human beings and computers in understanding real faces.

\subsection{Face Detection to Simulacra fake faces}
As mentioned in~\cite{takahashi15, abaci15}, we confirmed that face detectors trained with real faces could not detect simulacra fake faces. To accomplish this, we chose two types of face detectors, VJ~\cite{viola01} and CNN ~\cite{liu16}, and applied them to simulacra fake faces in an in-house created database~(see Section~\ref{sec:fakefacedb}). Next, the simulacra fake face detection accuracy was compared with that of images with and without real faces.

\subsection{Face Classification Trained with Fake faces}

Here, we discuss our attempt to classify real faces using CNN~\cite{krizhevsky12} trained with fake faces. However, it is important to note that real faces are not fed into a CNN classifier during the training period. In the testing phase, we attempted to verify whether the trained CNN could classify real faces and other objects. In this paper, to simplify the experiment, binary classification of fake faces and other objects was conducted. Let $y \in \{0, 1\}$ denote a class label including fake faces and other objects, $I$ denote an image from training set, $\theta$ denote trainable parameters, and loss function for the CNN $f(I, \theta, y_i)$ is written as:
\vspace{-3.0pt}

\begin{equation}
  \mathcal{L}_{cl}(f(I, \theta, y_i)) = -\log\left(\frac{\mathrm{e}^{f(I, \theta, y_i)}}{\sum_{j} \mathrm{e}^{f(I, \theta, y_j)}} \right)
\end{equation}


\subsection{Fake face Generation Trained with Fake faces}
In our experiment, we used GAN \cite{goodfellow14}to visualize the behavior of fake face images. The GAN architecture used in these experiments consists of a generator $G$ and a discriminator $D$. These networks are trained adversarially as standard GAN. The adversarial loss is defined in the ordinary manner:

\vspace{-3.0pt}
\begin{equation}
\mathcal{L}_{GAN} = \mathbb{E}_{\bm{x}\sim p_{data}(\bm{x})}[\log{D(\bm{x})}] \notag + \mathbb{E}_{\bm{z}\sim p_{\bm{z}}(\bm{z})}[1-\log(G(\bm{z}))] 
\label{eq:loss_gan}
\end{equation}

\section{Collection of Fake face Database}
\label{sec:fakefacedb}


\subsection{Fake faces Detected by Face Detectors}
To collect fake faces via face detectors, we classified the images in a large-scale database.
Databases like ImageNet~\cite{deng09} collect images of numerous kinds of objects such as cats and dogs. As a result of our preliminary experiment, we gathered images with labels that are frequently detected as faces and used those images in our training.
To avoid the overfitting specific objects, we also collected fake faces from Places365-Challenge~\cite{zhou17} using VJ and CNN face detectors.

As mentioned above, VJ detects faces by using thousands of dimensional handcrafted features, while CNN detect faces using millions of features. Thus, we regard VJ and CNN false positives as different objects. The images detected as faces naturally include real as well as fake faces. To remove the real faces from the detected images, we applied semantic segmentation~\cite{long15}.

We obtained the per-pixel labels of entire images using semantic segmentation. Since human faces often appear simultaneously with human bodies, semantic segmentation for an entire image results in better accuracy for human label predictions than for just the detected area. If the per-pixel label map in a detected area does not contain human labels, we regard the image in the detected area as a fake face.

Totally, 26,006 VJ fake faces, and 77,885 CNN fake faces were collected for face detector use.

\subsection{Fake faces in Simulacra and Pareidolia}
First, we retrieved images tagged with the word ``pareidolia'' from Flickr \cite{flickr} and selected 785 photos that showed simulacra phenomenon. Next, we downloaded 528 images from the site "WHAT THE FACE"~\cite{wtface}, which contains numerous simulacra images. In total, we collected 1,313 simulacra images. We then manually annotated the positions of the eyes and mouth. Finally, we cropped each image into a square in order to ensure that its width and height was two times the distance between the eyes and that the center of the eyes was located at (0.5$\times$width, 0.3$\times$height). When the cropped region protruded from the original image, the protruding regions were filled with the average color of the image. 

\section{Experiments}

We investigated computer-driven fake face understanding through image detection, classification, and generation. Here, we employed an in-house created fake face database (described in Section~\ref{sec:fakefacedb}). The settings and results of simulacra fake face detection, face classification, and generation are described in each subsection.

\vspace{-8.0pt}
\subsection{Analysis for Face detection to simulacra fake faces}
\subsubsection{Settings of face Detection to Simulacra Fake faces}
To validate CV methods for recognizing simulacra fake faces, we adapted face detectors to simulacra fake faces, real faces, and images without faces. More specifically, we used 1,313 simulacra fake faces from our fake face database, 50,000 real faces from CelebA,~\cite{liu15}, and 1,313 images without faces obtained from Places365-Challenge. To these images, we adapted VJ and CNN face detectors and then calculated the detection rate for simulacra fake faces, real faces, and the misdetection frequency.

\vspace{-2.0pt}
\subsubsection{Results and Discussion of Face Detection Using Simulacra Fake faces}

As shown in \tabref{tab:simures}, VJ and CNN face detectors have far lower detection accuracy for simulacra fake faces than is found for real faces. In particular, we found that VJ face detectors detect simulacra fake faces at the same frequency rate as they do for images that do not contain real faces. This result shows that VJ has the same level of accuracy for simulacra fake face detections as it has for misdetection accuracy, and that VJ considers simulacra fake faces to be extremely different from real faces. Moreover, even though DNN exceeds human performance, they still have low accuracy for recognizing simulacra fake faces. These results not only show that face detectors do not recognize simulacra fake faces as real faces, but that they also do not recognize them as false positive faces. Taken together, they also confirm that existing vision-based methods do not see similarities between simulacra fake faces and real faces.

\begin{table}[t]
\begin{center}
\caption{Detection rate for image queries each face detectors.}
\label{tab:simures}
\begin{tabular}{lccc}
\hline\noalign{\smallskip}
Method
& ~Simulacra fake faces~
& ~Real faces~
& ~Images without faces\\
\noalign{\smallskip}
\hline
\noalign{\smallskip}
VJ & 1.06 & 87.4 & 0.838\\
CNN & 7.09 & 86.6 & 1.83\\
\hline
\end{tabular}
\end{center}
\end{table}
\setlength{\tabcolsep}{1.4pt}

\begin{table}[t]
\begin{center}
\caption{Result of real face classification trained with each fake face type.}
\label{tab:clasi}
\begin{tabular}{lccc}
\hline\noalign{\smallskip}
measurements 
& \begin{tabular}{l}Trained with \\ VJ fake face \end{tabular} 
& \begin{tabular}{l} Trained with \\ CNN fake face \end{tabular}
& \begin{tabular}{l} Trained with \\ simulacra fake face \end{tabular} \\
\noalign{\smallskip}
\hline
\noalign{\smallskip}
Precision  & 0.991 & 0.987 & 0.877\\
Recall & 0.997 & 0.999 & 0.829\\
F-measure & 0.994 & 0.993 & 0.852\\
\hline
\end{tabular}
\end{center}
\end{table}
\setlength{\tabcolsep}{1.4pt}

\subsection{Analysis of Face Classifications Trained with Fake faces}

\subsubsection{Face Classification Settings.} 
In this analysis, we trained our classifier to catalogue fake faces as positives, non face objects as negatives and confirmed its ability to assign those classifications correctly during the testing phase.

To analyze current false positives by face detectors, we fine-tuned ResNet-50~\cite{he16} with our fake face database in the face classification stage. For negative samples, we collected non face images from MS COCO~\cite{lin14}.

We began by cropping images via bounding box annotations and then adapted them to the face detectors. Cropped images in which the face detectors did not detect faces were regarded as negative samples. When conducting testing, we used the face images from CelebA that were cropped by the CNN face detector as true data. Three experiments, VJ, CNN, and simulacra were conducted. Through all our experiments, the positive and negative images used in training were fake faces and non faces, respectively; whereas the positive and negative images used in testing were real faces and non faces, respectively. In the VJ, CNN, and simulacra experiments, the numbers of images used for training were 20,800, 20,800, and 1050 respectively, whereas the images used for testing were 5,206, 5,206, and 263 respectively, and the applied face detectors were VJ, CNN, and VJ, respectively.

Please note that in VJ experiments, images are converted to grayscale, and then input to the classifier.
In all experiments, each classifier was trained for 20 epochs.

\subsubsection{Face Classification Results and Discussion.} 
As shown in \tabref{tab:clasi}, classifiers trained with VJ/CNN fake faces  can successfully classify real faces as positives. This result suggests that face detectors perceive that fake faces have characteristics that are similar to real faces in feature space. Therefore, we could confirm that AI face detectors recognize fake faces in ways that are similar to those used to detect real faces. However, to human beings, the fake faces identified by face detectors (see \figref{fig:faces} (b) and (c)) are extremely different from real faces. This indicates that there is still a gap between humans and AI in face understanding.

Furthermore, \tabref{tab:clasi} shows that the classifier trained with simulacra fake faces is reasonably accurate, which tells us that it is capable of discerning some real-face features from simulacra fake faces. Therefore, we can conclude that a better understanding of simulacra fake faces could provide one of the keys to filling the gap between humans and AI in face understanding.

\subsection{Analysis for Fake face Generation Trained with Fake faces}

\subsubsection{Settings of Fake face Generation.}

In fake face generation, we use deep convolutional GAN (DCGAN) \cite{radford15} as our GAN model. The numbers of training images were 26,006, 77,885, and 1,313 for the VJ, CNN, and simulacra experiments, respectively. Due to the small number of simulacra images, we doubled the number of images to 2,616 by flipping them. We then trained our networks with the 2,616 simulacra images over 1,000 epochs. In all our experiments, the images were resized to 64 $\times$ 64 and then trained with DCGAN for 77,000 global steps in mini-batches of 100 data. 

\subsubsection{Results and Discussion of Fake face Generation.}

The results of fake face generation are shown in \figref{fig:generated}. Note that we refer to the results generated by our GAN trained with VJ fake faces as VJ results. GAN trained with CNN and simulacra fake faces are referred to as CNN results and simulacra results, respectively. The images shown in \figref{fig:generated} (a) and (b) are VJ the and CNN results. Here, we can see that those images have something resembling eyes and a mouth as well as facial contours. These results suggest that VJ fake faces and CNN are close to real faces in feature space, whereas most of them do not look like real human faces. Images generated by the generator trained with simulacra fake faces (\figref{fig:generated}) show that the eyes and mouth are synthesized more clearly than the VJ/CNN results. Furthermore, in contrast to the VJ/CNN results, the images in simulacra result have no contours. By comparing the false positives of vision-based methods (VJ/CNN) with false positives of humans (simulacra), we find that while humans focus strongly on the eyes and mouths, vision-based methods not only focus on the eyes and mouths, they also note contours when detecting faces.

\iffigure
\begin{figure}[t]
  \begin{center}
    \includegraphics[clip,scale=0.45]{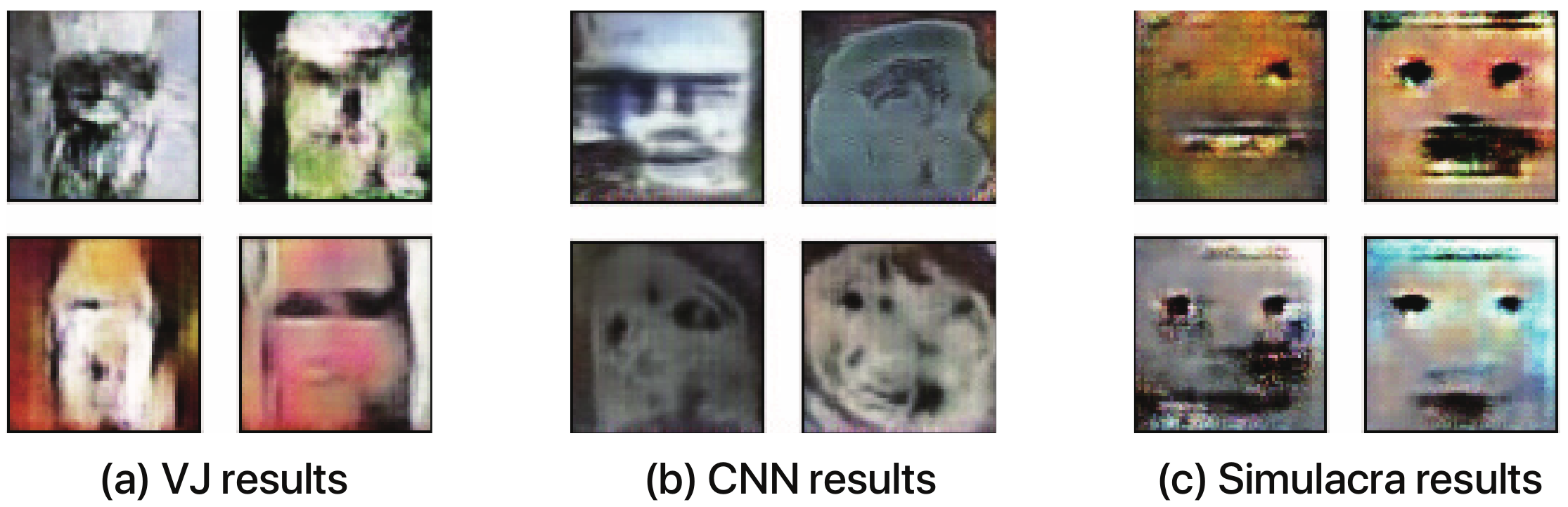}
    \vspace{-8.0pt}\caption{Results of fake face generation. (a) and (b) are trained by VJ and CNN images. (c) is trained with simulacra fake faces. \vspace{-5.0pt} \label{fig:generated}}
  \end{center}
\end{figure}
\fi

\section{Conclusion}
In this paper, we surveyed the face understanding of AI algorithms by creating a novel fake face database that includes AI (face detector~\cite{viola01, liu16} and human false positives (i.e. pareidolia~\cite{liu14}). We also conducted face detection and generation experiments with our fake face database and found a level of suggestive knowledge that indicates the continuing existence of a gap between the capabilities of recent vision-based face recognition algorithms and human-level performance. On a positive note, however, we also obtained knowledge that will advance the progress of face-understanding models.

\section*{Acknowledgments}

This study was granted in part by the Strategic Basic Research Program ACCEL of the  Japan Science and Technology Agency (JPMJAC1602).  Shigeo Morishima was supported by a Grant-in-Aid from Waseda Institute of Advanced Science and Engineering. We have had the support and encouragement of cvpaper.challenege group.

\bibliographystyle{splncs}
\bibliography{reference}
\end{document}